%% file: egpaper_final.tex
\documentclass[10pt,twocolumn,letterpaper]{article}

\usepackage{cvpr}
\usepackage{times}
\usepackage{epsfig}
\usepackage{graphicx}
\usepackage{amsmath}
\usepackage{amssymb}
\usepackage{mathrsfs}
\usepackage{makecell}
\usepackage{multirow}

\usepackage[breaklinks=true,bookmarks=false]{hyperref}

\newcommand\blfootnote[1]{%
  \begingroup
  \renewcommand\thefootnote{}\footnote{#1}%
  \addtocounter{footnote}{-1}%
  \endgroup
}

\cvprfinalcopy 

\def\cvprPaperID{****} 

\begin{document}

\title{Efficient Video Understanding via Layered Multi Frame-Rate Analysis}


\author{Ziyao Tang\thanks{Equal contribution}\qquad Yongxi Lu\footnotemark[1]\qquad Tara Javidi\\University of California, San Diego}

\maketitle
\blfootnote{arxiv preprint, work in progress.}

\begin{abstract}
   One of the greatest challenges in the design of a real-time perception system for autonomous driving vehicles and drones is the conflicting requirement of safety (high prediction accuracy) and efficiency. Traditional approaches use a single frame rate for the entire system. Motivated by the observation that the lack of robustness against environmental factors is the major weakness of compact ConvNet architectures, we propose a dual frame-rate system that brings in the best of both worlds: A modulator stream that executes an expensive models robust to environmental factors at a low frame rate to extract slowly changing features describing the environment, and a prediction stream that executes a light-weight model at real-time to extract transient signals that describes particularities of the current frame. The advantage of our design is validated by our extensive empirical study, showing that our solution leads to consistent improvements using a variety of backbone architecture choice and input resolutions. These findings suggest multiple frame-rate systems as a promising direction in designing efficient perception for autonomous agents. 
\end{abstract}

\begin{figure*}
    \centering
    \includegraphics[width=\linewidth]{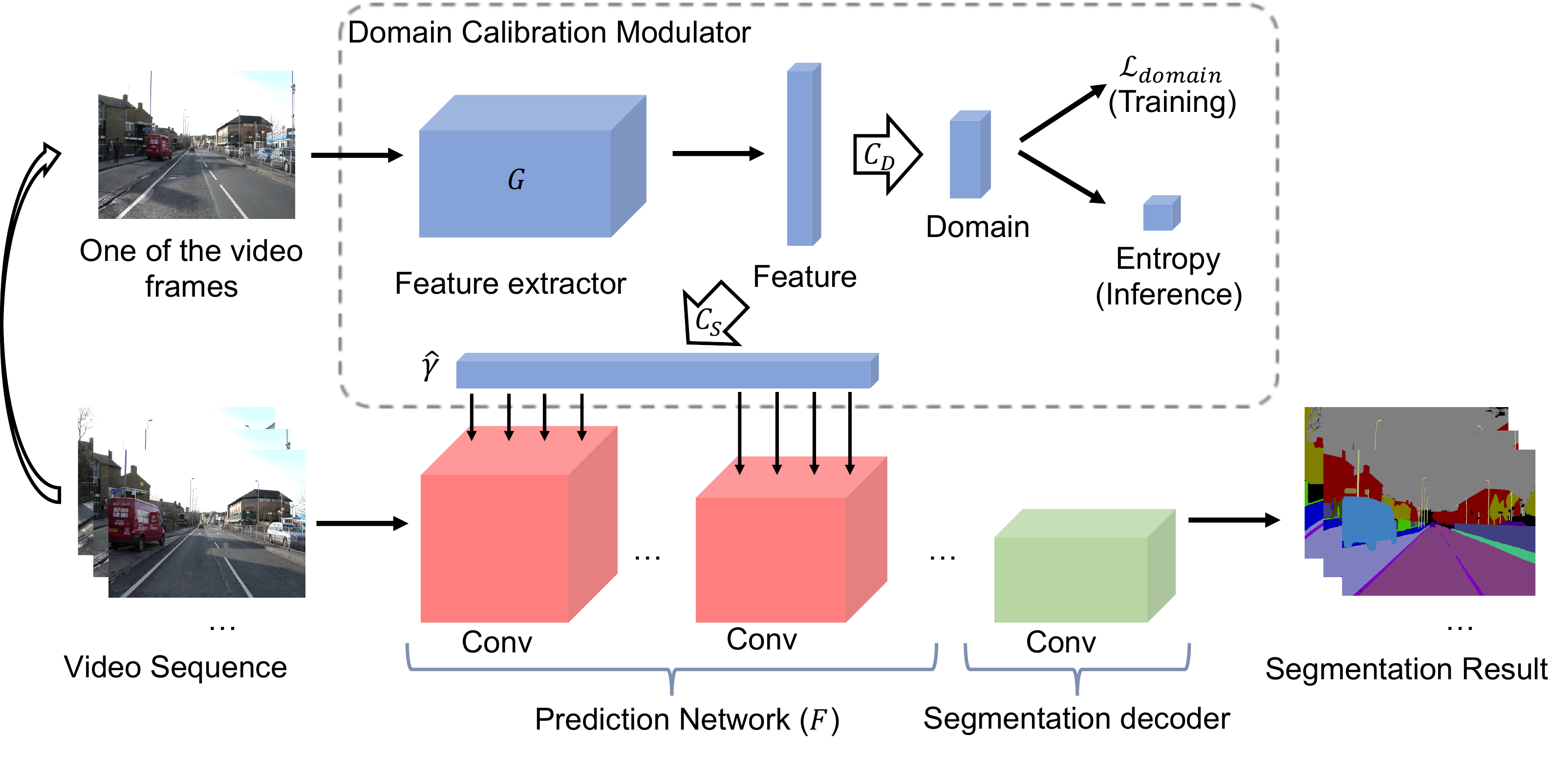}
    \caption{The pipeline of our proposed framework, using semantic segmentation as an example. }
    \label{fig:pipeline}
\end{figure*} 

\input{tex/Introduction.tex}
\input{tex/RelatedWork.tex}

\input{tex/Approach.tex}

\input{tex/Experiments.tex}

\input{tex/Conclusion.tex}
{\small
\bibliographystyle{ieee}
\bibliography{egbib}
}

\end{document}

%% file: tex/Introduction.tex
\section{Introduction}

Models based on deep convolutional neural networks (ConvNets) have become the de facto standards in a wide variety of computer vision tasks, such as image classification \cite{Alpher10, Alpher11, Alpher13, Alpher13a}, object detection \cite{Alpher19, Alpher19a, Alpher19b}, semantic segmentation \cite{Alpher12, Alpher12a, Alpher12b} and instance segmentation \cite{Alpher19c, Alpher19d, Alpher19e}. While deep ConvNets have shown excellent performance and in many cases greatly simplified the learning framework, modern neural networks require expensive computations and are resource intensive. The most successful ConvNet architectures for vision applications, such as VGG \cite{Alpher11}, ResNet \cite{Alpher13} and DenseNet \cite{Alpher13b} introduce high latency and high energy consumption at run time. This has been an obstacle for their adoption in low-latency and low-energy applications with embedded hardware. 

There has been a growing interest in designing compact architectures, for example through architecture search \cite{Zoph2016NeuralAS,Lu2017FullyAdaptiveFS,Pham2018EfficientNA,Zoph_2018_CVPR,Zhong_2018_CVPR,Liu_2018_ECCV}, network compression \cite{He_2018_ECCV,Chen_2018_ECCV,Peng_2018_ECCV} and manual architecture design \cite{Howard2017MobileNetsEC,Zhang2017ShuffleNetAE,Alpher14}. The prior works, however, by focusing on single-frame tasks, have not utilized the rich correlation between frames often present in the sequence of frames of given stream in most practical applications. Spatially adaptive processing \cite{Lu2016AdaptiveOD,Figurnov_2017_CVPR} and dynamic layer dropping \cite{Wu_2018_CVPR,Wang2017SkipNetLD,Veit2017ConvolutionalNW} have been proposed to determine sample-dependent computational architecture on-the-fly. Such a design usually leads to significant reduction of the average computations required to achieve the same accuracy level. However, a dynamic architecture inevitably introduces extra overhead in the architecture expansion process. The resultant algorithms can be practically less efficient in modern accelerators since the architecture cannot benefit from parallel processing as easily. 

In video object detection, temporal consistency of labels have been used to improve accuracy while reducing complexity \cite{Bertasius_2018_ECCV}. However, training of such a system requires consecutive dense labels of frames, which is time consuming and tedious to collect. The resultant model is also arguably more sensitive to labeling errors. At test time, its performance could be affected by common failures of the hardware and software systems in embedded/mobile applications, such as frame dropping and unstable frame rate. 

In this work, we propose an alternative framework to complement the existing methods. In video-based applications, much of the intra-class variations can be attributable to changes in persistent environmental factors, such as lighting conditions, scenes, weather or just the general ``style'' of a particular geographical destination. This is similar to the so-called ``domain-shift'' \cite{Alpher01, Alpher02a, Alpher02d,Alpher03,Alpher04a,Alpher05a,Alpher05b} in domain adaptation literature, but in our case the domain-shift is temporal. A real-time compact ConvNet architecture is not robust to such domain-shifts as it does not have the same level of feature extraction redundancies in expensive architectures. However, deploying a large model at fast frame-rate is difficult and resource intensive. Our main observation is that the persistent factors that cause the temporal domain-shifts change slowly: In many applications a set of consecutive frames at the number of thousands (only roughly 30 seconds at 30 FPS) could look more or less the same, save for some small details. This suggests the following two-stream solution: The upper stream processes information at a slower frame rate using an expensive model to extract features that describes the environment. The extracted features is then used to modulate a real-time lower stream to make it more robust. Our main finding in this paper is that such a design can yield consistent gains over baselines at only negligible increase in complexity. 

The contributions of this work are as follows. 

\begin{itemize}
\item We propose a novel two-stream architecture for fast video understanding. This system demonstrates consistent improvement in accuracy with negligible increase in complexity. 
\item We perform extensive empirical studies on semantic segmentation tasks, showing that the proposed system is robust to different input resolutions and the choice of neural network architectures. 
\item Through ablation studies, we confirm that the improvements result from correctly modeling the environment. It cannot be attributable to trivially increasing parameters or complexity in the models. 
\item We analyze the effects of incorrect prediction of the environment and propose a simple heuristic that detects such failure cases. This leads to a simple rejection sampling strategy which further improves the robustness of our solution. 
\end{itemize}

The rest of the paper is organized as follows. Section \ref{sec:related} summarizes the related works. We then introduces our methods in Section \ref{sec:method}. Our experiments and ablation studies are presented in Section \ref{sec:exp}. Section \ref{sec:conclusion} concludes our paper and points to future directions. 

%% file: tex/RelatedWork.tex
\section{Related Work}
\label{sec:related}

\noindent\textbf{Fast Neural Networks} Architecture search \cite{Zoph2016NeuralAS,Lu2017FullyAdaptiveFS,Pham2018EfficientNA,Zoph_2018_CVPR,Zhong_2018_CVPR,Liu_2018_ECCV}, network compression \cite{He_2018_ECCV,Chen_2018_ECCV,Peng_2018_ECCV} and novel manual design \cite{Howard2017MobileNetsEC,Zhang2017ShuffleNetAE,Alpher14} are popular methods in designing fast neural network architectures. These methods can find efficient architectures, but by focusing on single-frames they cannot utilize the strong correlation between nearby frames which are usually present in video understanding applications. Spatially adaptive processing \cite{Lu2016AdaptiveOD,Figurnov_2017_CVPR} and dynamic layer dropping \cite{Wu_2018_CVPR,Wang2017SkipNetLD,Veit2017ConvolutionalNW} achieves efficient execution through conditional execution on part of the model or image, but these methods introduce additional overheads in decision makings and are less efficient in practice. Our solution is a simpler static architecture. Our method is also related to \cite{Bertasius_2018_ECCV}. Similar to our method, they utilize the correlation between frames to achieve efficient processing. However, they focus on modeling the temporal consistency of labels. Our method focuses on separating environment modeling and per-frame modeling in a dual frame-rate system. 

\noindent\textbf{Domain Adaptation and Network Modulation}
Domain adaptation aims at improving generalization of models across different domains \cite{Alpher01, Alpher02a, Alpher02b, Alpher02c, Alpher02d,Alpher03,Alpher04a,Alpher04b,Alpher05a,Alpher05b,Alpher05c}. ``Environment factors'' we model in this work can be seen as a particular way to partition images into different domains. However, our work does not aim at improving generalization per se. Instead, it acknowledges that a light-weight model is unlikely to generalize well for all domains. To battle domain-shifts, an expensive model extracts robust features to augment the light-weight model through modulation. Our modulation algorithm is similar to Conditional Batch Normalization (CBN) \cite{Alpher06, Alpher07, Alpher08}, which has shown to be effective in controlling domain distributions, in applications such as style transfer and question answering. In our case, CBN is used to influence the feature distribution of the prediction network. A similar modulation pipeline is introduced in \cite{Yang_2018_CVPR}. Similar to ours, this work modulates the feature extraction process at the current frame using features from a different frame. However, their approach is designed for one-shot learning. Our method addresses a different problem. 

\noindent\textbf{Attention Mechanisms}
Self-attention mechanisms \cite{Wang2017ResidualAN,Hu_2018_CVPR} leads to better generalization by learning automatic recalibration of features. Our method similarly calibrates the features of the light-weight network to improve its generalization. Unlike a self-attention mechanism, the features used to perform calibration in our case are not from the current image or model. They are extracted from a nearby frame using a different model. 

\noindent\textbf{Semantic Segmentation}
Our method is evaluated in a semantic segmentation task using the classical FCN decoder \cite{Alpher12} as our baseline. Our contribution is complementary to state-of-the-art methods in segmentation, such as \cite{Alpher12, Alpher12a, Alpher12b}. We expect to see similar accuracy gains from using our method on top of the latest segmentation models.

%% file: tex/Approach.tex
\section{Methods}
\label{sec:method}

In this section we introduce our proposed system: A light-weight per-frame prediction model guided by a heavy-lifting domain calibration modulator (DCM). The per-frame prediction model is deployed at high frame-rate to catch all the subtle changes in each frame. It is trained to output the predicted labels (e.g. classification, detection or semantic segmentation etc.). The heavy-lifting domain calibration modulator (DCM) is deployed at a much lower frame-rate to reduce latency and power consumption, but it predicts reliable features which are fed into the prediction model through a ``modulation'' mechanism to improve task prediction robustness. Interestingly, this proposed framework consistently improve the test accuracy while adding only negligible amount of complexity per frame. We will present our empirical findings in the next section. The current section presents the basic structures of our framework.

\subsection{Two-Stream Architecture}

The goal of the proposed framework is to learn a function $\hat{y}(x; \theta)$ so as to minimize $\mathbb{E}[\mathcal{L}_{D} (y, \hat{y}(x; \theta))]$, where $(x,y)$ are pairs of images and the associated task labels. The data consists of a number of domains $\{D_1, \cdots, D_k\}$. Images within the same domain are visually similar but could have very different labels. 

In applications that use streaming inputs, short-term contiguous sequence of frames naturally falls into the same ``domain''. In view of this structure, to reduce complexity while improving prediction accuracy we decouple the predictor $\hat{y}$ into two components: A domain calibration modulator (DCM) and a prediction model. Figure \ref{fig:pipeline} illustrates the framework using semantic segmentation task as an example. The DCM model extracts features shared across images in the same domain using an arbitrary sample from the domain. It then uses this feature to predict a set of calibration parameters $\hat{\gamma}_c$. The prediction model extracts fine-grained features from each frame. In the interest of fast processing, the prediction model is compact and as a result its features are less robust. This is compensated by a modulation mechanism which applies the calibration parameters to the prediction model. This process ``calibrates'' the inaccurate features extracted from the prediction model and turn them into robust features for the current domain.




\begin{figure}
    \centering
    \includegraphics[width=\linewidth]{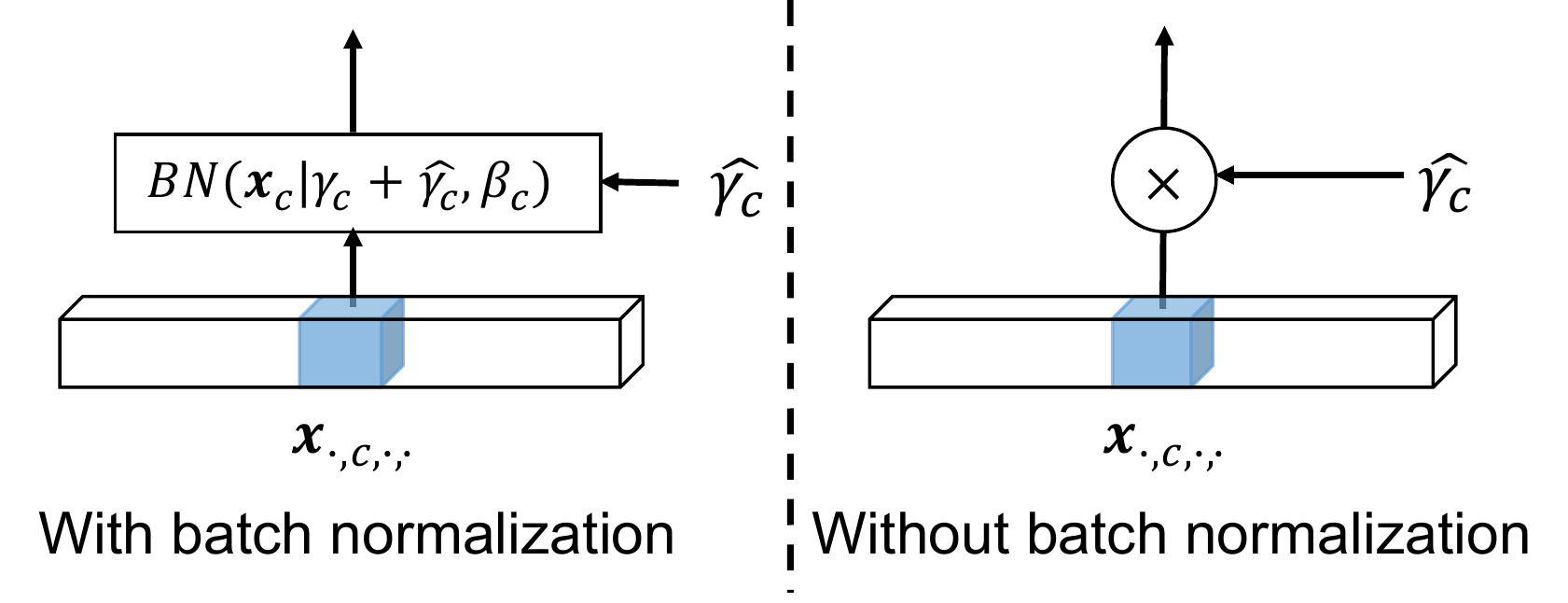}
    \caption{Proposed modulator design in networks without BN layers (left) and with BN layers (right). }
    \label{fig:modulator}
\end{figure} 

\subsection{Domain Calibration Modulator}
\label{sec:dcm}

There are two important choices in the design of a DCM architecture, namely how are calibration features extracted and how can it be used to modulate the prediction network. 

\textbf{How are features extracted?} The intermediate and output features extracted by a ConvNet architecture are multi-scale and dense. However, in DCM the features are out-of-sync with the current frame. Spatial details extracted should be suppressed or modulation based on it will be noisy and confusing since it is describing the incorrect contents. For this reason, we propose to use global average pooling from the last convolution layer to perform calibration feature extraction. The features extracted in this way are global descriptors, making them more suitable and stable for describing the environmental information. 

\textbf{How to modulate the prediction network using calibration features?} The main consideration in this design choice is the trade-off between complexity and accuracy. If the modulation function is introducing large extra complexity, it will defeat the purpose of our design. We propose a modulation mechanism based on channel-wise scaling. This mechanism is orders of magnitude lighter than a convolution operator, yet it is shown to be very effective in other related controlling and re-calibrating the distribution of features \cite{Hu_2018_CVPR,Alpher06,Alpher07,Alpher08,Yang_2018_CVPR}. With this mechanism, our method attenuates or amplifies the contribution of feature channels based on the decoding of the environment by the DCM network. The DCM in effect adaptively selects the most useful features from the prediction network for the current domain. In our design, the modulation scalings are produced by a simple linear transformation of the extracted calibration features (from global average pooling). 

The final design can be described in simple formulas. As shown in Figure \ref{fig:pipeline}, the proposed method first extracts features from a sample frame, then predict per-channel scales through a linear transformation

\begin{equation}
\label{eq:3}
    \hat{\gamma} = C_{S}(G(\mathbf{x}))
\end{equation}

The per-channel scales are applied to the prediction network through element-wise modulation with a few selected intermediate layers. For prediction networks without Batch Normalization, the modulation mechanism is applied after every convolution layer. It can be written as

\begin{equation}
\label{eq:1}
    F(\mathbf{x}_{i,c,h,w}|\hat{\gamma_{c}}) = \hat{\gamma_{c}}\mathbf{x}_{\cdot,c,\cdot,\cdot} 
\end{equation}

For prediction networks with Batch Normalization (BN), these BN layers are replaced by

\begin{equation}
\label{eq:2}
    BN(\mathbf{x}_{i,c,h,w}|\gamma_{c}, \hat{\gamma_{c}}, \beta_{c}) = (\gamma_{c} + \hat{\gamma}_{c})\frac{\mathbf{x}_{\cdot,c,\cdot,\cdot} - E[\mathbf{x}_{\cdot,c,\cdot,\cdot}]}{Var[\mathbf{x}_{\cdot,c,\cdot,\cdot}] + \epsilon} + \beta_{c}
\end{equation}
where $\gamma_{c}$ is the original scale factor of the BN layer. In this case, the predicted scale is the residual of the modified BN scale factor. Using a residual scale is particularly useful when a pretrained model is used, as this avoid destructing the original BN parameters. The differences in design with and without BN are summarized in Figure \ref{fig:modulator}.

\begin{figure}
    \centering
    \includegraphics[width=\linewidth]{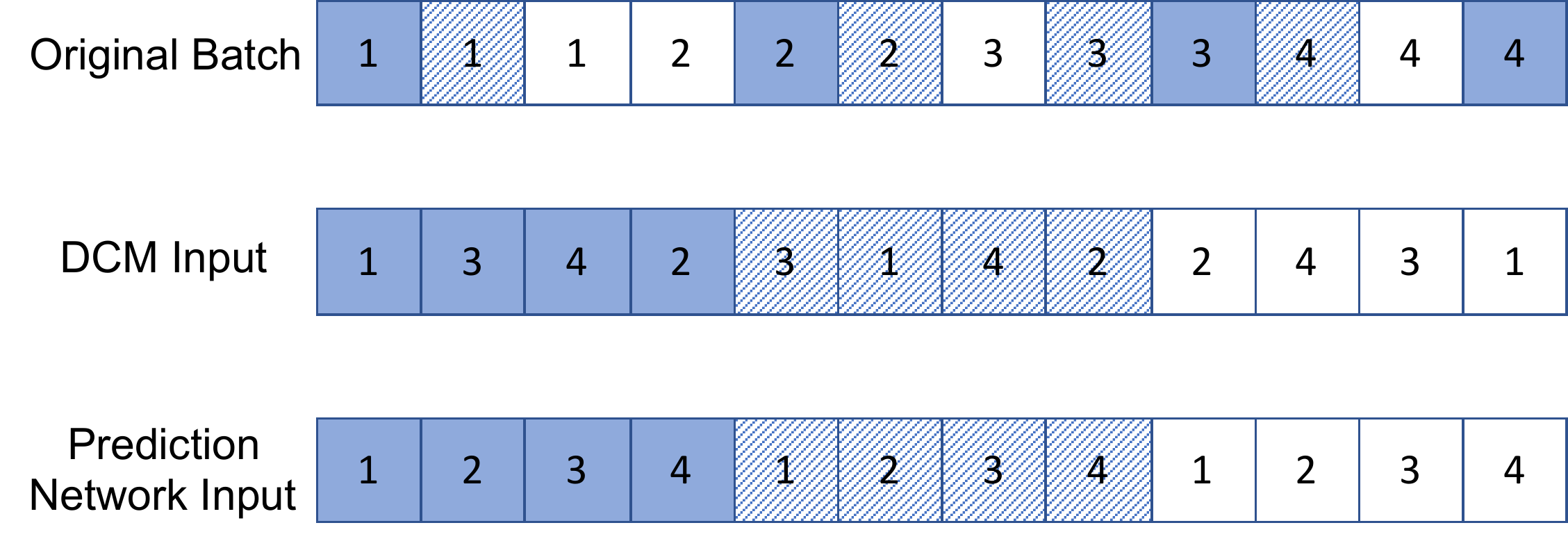}
    \caption{The sampling of training batches. }
    \label{fig:batch}
\end{figure} 

\subsection{Model Training and Loss Functions}
\label{sec:training}

Training of our system requires image pairs from the same domains. In the applications we consider, the datasets are separated into video clips, each containing frames taken at similar time and locations. In both the training and testing procedures, these groups are considered ``domains''. This partition of data does not guarantee that all similar image pairs are classified into the same domain. But it suffices in providing image pairs that are likely to be visually similar. Based on this domain partition, our training procedure is essentially a two-step sampling process. It fills a training mini-batch in the following fashion (also see Figure \ref{fig:batch}):

\begin{enumerate}
\item Sample a set of images from the entire datasets. These images are used as input to the DCM. Their domain ids (video clip names) are recorded.
\item Group sampled images in step 1 according to domain ids. Re-shuffle samples within each group. The images are used as input to the prediction network using this new ordering. 
\end{enumerate}

The system can be trained end-to-end via a loss function that penalizes incorrect predictions at the output head of the prediction network. The DCM model is trained jointly with the prediction network as our proposed modulation mechanism is differentiable. 

Features extracted for calibration should intuitively be similar if they are from two images in the same domain. To regularize the calibration features, in our framework, if there are $k$ video clips, we train a $k$-way domain classifier using a simple linear layer from calibration features (shown as $C_D$ in Figure \ref{fig:pipeline}). The video clip id is the domain label in this case. The entropy loss serves to regularize  the calibration features. In addition, the entropy of the domain classifier prediction can also indicate the quality of the extracted domain features. This results in a more robust testing procedure for unexpected changes in the environment. Details of our testing procedure can be found in Section \ref{sec:inference}. 

We denote the per-frame prediction network as $F$, and the domain label as $d$. The total loss for our framework is the sum of the task loss and the domain loss presented. The total loss function is
\begin{equation}
\label{eq:7}
\begin{aligned}
    \mathcal{L}_{total} &= \mathcal{L}_{domain} + \mathcal{L}_{task} \\
    & = \frac{1}{N}\sum_{i,j\in[1, N]}\{\mathcal{L}(C_{D}(G(\mathbf{x}_{j}^{d})), d) \\
    & \ \ \ \ + \mathcal{L}(F(\mathbf{x}_{i}^{d}), \mathbf{y}_{i}^{d};\hat{\gamma})\} \\
    & = \frac{1}{N}\sum_{i,j\in[1, N]}\{\mathcal{L}(C_{D}(G(\mathbf{x}_{j}^{d})), d) \\
    & \ \ \ \ + \mathcal{L}(F(\mathbf{x}_{i}^{d}), \mathbf{y}_{i}^{d};C_{S}(G(\mathbf{x}_{j}^{d})))\}
\end{aligned}
\end{equation}

\subsection{Model Inference}
\label{sec:inference}

A practical application of such a system is to use the first few frames at any operating sessions to extract ``calibration'' features using the DCM model. Assuming a slowly changing environment, the extracted calibration features are then re-used by the prediction model until the end of the particular deployment session. However, by the design of our training procedure as described in Section \ref{sec:training}, our system should work equally well with out-of-order stream of inputs. It does not rely on the particular ordering of frames as long the inputs to the DCM and the prediction network are in the same domain. To confirm this conjecture, in our experiments we evaluate our system in both cases: (1) Feeding the DCM model with randomly selected inputs from the same domain (2) Feed the DCM model with the first few frames within the video clip. In both cases, our system demonstrates similarly strong improvements over the baselines methods. 


An interesting observation from our empirical study is that if the domain prediction has high entropy, it strongly indicates a reduction in final test accuracy using this particular image for calibration. This could be caused by model failure, changes in environmental factors or a frame being not representational for a particular domain. In view of this, we propose a robust version of our naive testing procedure. In this alternative entropy-based testing, we perform a rejection sampling. If an incoming candidate input to DCM results in high entropy in domain prediction, we disqualify this candidate and sample another example. Our testing procedure can further improve average and worst-case performance of our system while using only 1-2 additional samples for the DCM, which is only a small increase in complexity. 

\noindent\textbf{Complexity at Inference} We conclude this section with details in the computation of complexity of our testing procedure. Assuming that the DCM needs to process $K$ images before extracting the final calibration features and a particular operation session (video clip) with $N$ frames. Denote the computational cost per frame as $M_{Pred}$ for the prediction network and $M_{DCM}$ for the DCM. The average computational cost per frame ($M_{frame}$) is given by
\begin{equation}
\label{eq:8}
\begin{aligned}
     M_{frame} = M_{Seg} + \frac{K}{N}M_{DCM}
\end{aligned}
\end{equation}

In practical applications, $K\ll N$ in Equation \ref{eq:8}. In such cases, our framework does not bring in much increase in average complexity. In our experiment sections, the complexity metrics reported are numbers of multiply-adds.

%% file: tex/Experiments.tex
\section{Experiments}
\label{sec:exp}

To validate our design and test its applicability in real-world scenarios, we evaluate our framework on CamVid \cite{Alpher30b, Alpher30a} dataset.  We test the robustness of our approach under a variety of combinations of popular ConvNet architectures. In all test cases we observe a consistent accuracy gain with negligible increase in complexity against baseline methods. This suggests our method can be applied in a wide variety of settings. Through extensive ablation studies, we conclude that the gain from our approach is not a trivial result of naively adding parameters. Nor does it result from learning the dataset average. Our analysis also shows that the proposed rejection sampling using entropy further improves the robustness of the system. 

In the remainder of this section we will first discuss the choice of the dataset and data preprocessing details. We then discuss the details of our training and testing procedures and the exact modulation mechanisms used for different architectures and problem setups. It is followed by our main results on a variety of architectures, comparing our results against baselines available in the literature and our own replications. This section is concluded with ablation studies that sheds lights on how our system improves the performance and its most salient failure modes. 

\subsection{Dataset and Preprocessing}

\textbf{CamVid \cite{Alpher30b, Alpher30a}} is a segmentation dataset with 11 ground truth classes including sky, building, pole, road, pavement, tree, sign-symbol, fence, car, pedestrian, bicyclist. The images are all taken at street-level. It consists of four video sequences taken in different locations and times. We use the same training, testing and validation split as in \cite{Alpher16}. We use the training split for training and the testing split for evaluation.   

This dataset is ideally suited for evaluating our method: (1) It is separated into a few videos, ideally simulating the scenario in which environmental factors are persistent within each video clip. (2) It has full annotations of every frames of the videos, making it suitable for performing extensive ablation studies of our proposed framework. While there are other large-scale datasets in similar applications, their configurations are not suitable for testing our particular framework. However, conclusions drawn from CamVid should still be informative for future works on large-scale evaluations. 

Baseline methods \cite{Alpher16, Alpher17} processes images at $360\times 480$ and $720\times 960$. In our evaluations both settings are tested. In training time, we apply random crops of $352\times 352$ for $360\times 480$ images, and $704\times 704$ for $720\times 960$ images, respectively. Random horizontal flip is used as in standard practices. Inputs to DCM are scaled to $224 \times 224$.



\begin{figure}
    \centering
    \includegraphics[width=\linewidth]{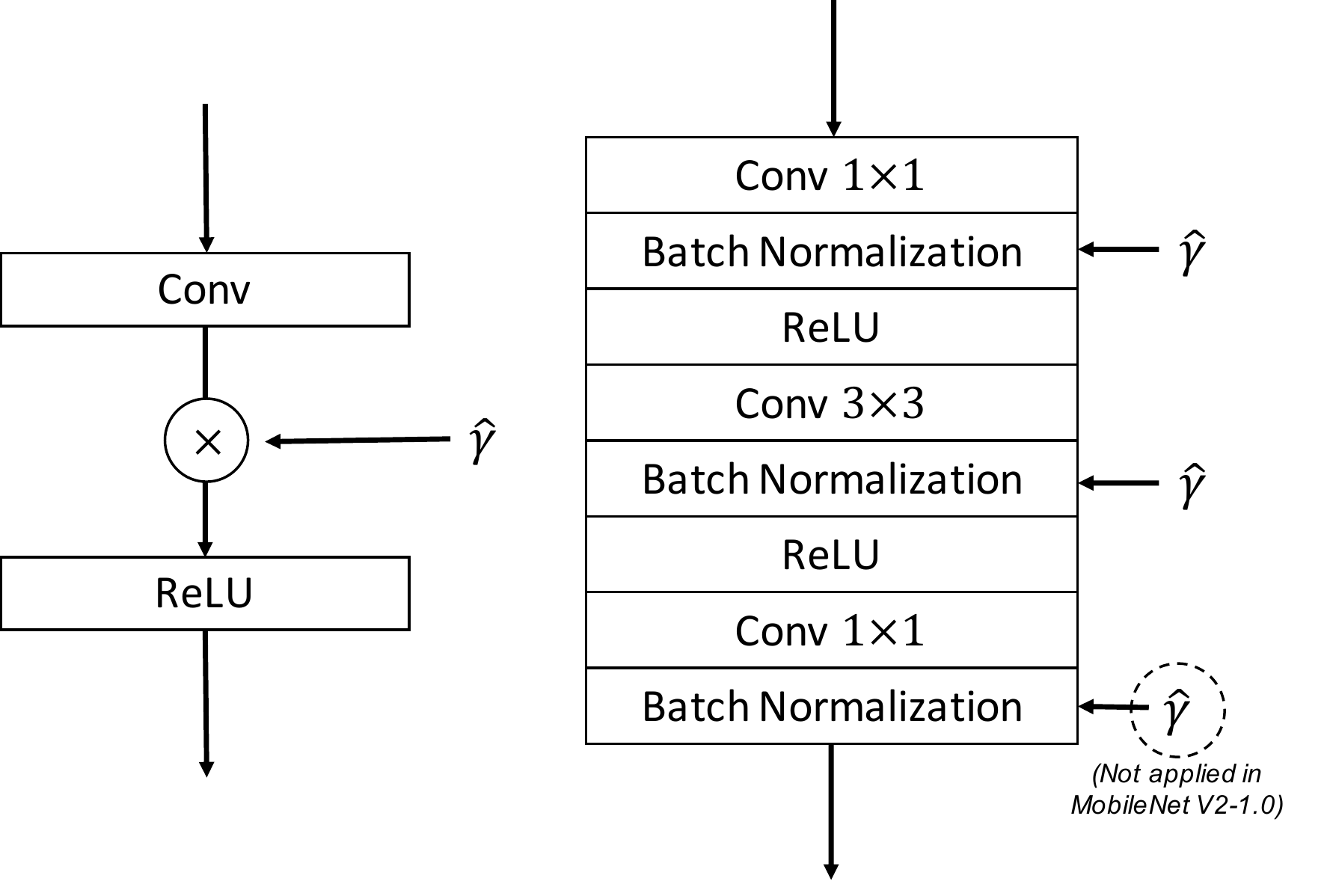}
    \caption{The modified convolution layer from AlexNet (left) and modified inverted residual unit from MobileNet V2 (right). }
    \label{fig:unit}
\end{figure}

\begin{table*}[t]
\begin{center}
\begin{tabular}{l|cc|lcc}
\hline
Backbone & Img. Size & Pretrained & mIOU($\%$) & Eval Img. & Mult-Adds / Frame \\ 
\hline
AlexNet(baseline) & & & $41.2$ & $-$ & $16.02G$ \\
AlexNet+DCM(Ours) & $360\times 480$ & $\times$ & $44.1(\pm0.42)$ & $2$ & $16.05G$\\
AlexNet+DCM+Entropy(Ours) & & & $\mathbf{44.4}(\pm0.43)$ & $4.55$ & $16.10G$\\
\hline
AlexNet(baseline) & & & $50.6$ & $-$ &$16.02G$ \\
AlexNet+DCM(Ours) & $360\times 480$ & \checkmark & $53.0(\pm0.18)$ & $2$ & $16.05G$\\
AlexNet+DCM+Entropy(Ours) & & & $\mathbf{53.1}(\pm0.15)$ & $3.55$ & $16.08G$\\
\hline
AlexNet(baseline) & \multirow{4}*{$720\times 960$} & \multirow{4}*{$\times$} & $44.0$ \cite{Alpher16} & $-$ & $54.79G$ \\
AlexNet(baseline, Ours) & & & $48.7$ & $-$ & $54.79G$ \\
AlexNet+DCM(Ours) & & & $51.7(\pm0.75)$ & $2$ & $54.82G$\\
AlexNet+DCM+Entropy(Ours) & & & $\mathbf{51.8}(\pm0.90)$ & $3.75$ & $54.89G$\\
\hline
AlexNet(baseline) & \multirow{4}*{$720\times 960$} & \multirow{4}*{\checkmark} & $57.4$ \cite{Alpher16} & $-$ & $54.79G$ \\
AlexNet(baseline, Ours) & & & $57.2$ & $-$ & $54.79G$ \\
AlexNet+DCM(Ours) & & & $60.8(\pm0.38)$ & $2$ & $54.82G$ \\
AlexNet+DCM+Entropy(Ours) & & & $\mathbf{60.9}(\pm0.37)$ & $3.1$ & $54.84G$\\ 
\hline
\end{tabular}
\end{center}
\caption{Result for AlexNet-FCN32s experiment. Eval Img. denotes to average number of images the DCM processed.}
\label{tab:a}
\end{table*}

\begin{table*}[t]
\begin{center}
\begin{tabular}{l|cc|lcc}
\hline
Backbone & Img. Size & Pretrained & mIOU($\%$) & Eval Img. & Mult-Adds / Frame \\ 
\hline
MobileNet-1.0(baseline) & & & $58.4$ & $-$ & $2747.49M$ \\
MobileNet-1.0+DCM(Ours) & $360\times 480$ & \checkmark & $59.2(\pm0.70)$ & $2$ & $2750.38M$\\
MobileNet-1.0+DCM+Entropy(Ours) & & & $\mathbf{59.6}(\pm0.08)$ & $2.45$ & $2751.04M$\\
\hline
MobileNet-0.75(baseline) & & & $58.1$ & $-$ & $1980.49M$ \\
MobileNet-0.75+DCM(Ours) & $360\times 480$ & \checkmark & $59.4(\pm0.23)$ & $2$ & $1983.36M$\\
MobileNet-0.75+DCM+Entropy(Ours) & & & $\mathbf{59.5}(\pm0.16)$ & $2.35$ & $1983.87M$\\
\hline
MobileNet-0.5(baseline) & & & $54.4$ & $-$ & $1069.92M$ \\
MobileNet-0.5+DCM(Ours) & $360\times 480$ & \checkmark & $56.6(\pm0.40)$ & $2$ & $1072.75M$\\
MobileNet-0.5+DCM+Entropy(Ours) & & & $\mathbf{56.9}(\pm0.07)$ & $3.15$ & $1074.37M$\\
\hline
MobileNet-0.35(baseline) & & & $51.8$ & $-$ & $770.34M$ \\
MobileNet-0.35+DCM(Ours) & $360\times 480$ & \checkmark & $52.0(\pm1.92)$ & $2$ & $773.14M$ \\
MobileNet-0.35+DCM+Entropy(Ours) & & & $\mathbf{53.0}(\pm0.21)$ & $2.5$ & $773.85M$\\ 
\hline
\end{tabular}
\end{center}
\caption{Result for MobileNet-FCN32s experiment. Eval Img. denotes to average number of images the DCM processed.}
\label{tab:b}
\end{table*}

\subsection{Network Architectures}

In experiments on CamVid, the DCM backbones are either ResNet50 \cite{Alpher13} or MobileNet V2-1.0 \cite{Alpher14}. We perform global average pooling at the top convolution layer to extract calibration features, in a manner described in Section \ref{sec:dcm} and Figure \ref{fig:pipeline}. 

The prediction networks use fully convolution networks (FCNs) \cite{Alpher12} decoders. The decoders follow the FCN32s configuration. To test performance of compact networks, the backbone architectures are AlexNet \cite{Alpher10} and MobileNet V2 with different width factor. For AlexNet backbones, channel-wise modulation is applied after every convolution layers. For MobileNet V2 backbones, by default DCN modulates every BN layers in the inverted residual blocks. However, for MobileNet V2-1.0 we omit modulation for the last convolution layer in each inverted residual block. Figure \ref{fig:unit} summarizes the modifications. 

\subsection{Training and Testing Procedures}

\textbf{Training Procedure and Hyperparameters}
We train all models with a batch size of $12$ on $3$ Nvidia GTX 1080Ti GPUs. We use Adam optimizer \cite{Alpher18} with weight decay of $0.0005$. Unless specified otherwise, we train for $600$ epochs. When the DCM model uses ResNet50 backbones, we use a step-wise learning rate schedule with initial learning rate of $0.0001$ and reduces it to $0.00001$ at epoch $400$. When the DCM model uses MobileNet v2-1.0 backbone, the initial learning rate is $0.00005$ with reduction to $0.000005$. The training mini-batches are sampled according to the procedure described in Section \ref{sec:training}.

The scale prediction layer of the DCM ($C_{s}$) is a fully connected layer initialized with all-zero weights and the bias is set to $1$. This ensure that the initial output of the DCM is always $1$. In MobileNet based prediction networks, if a pretrained model is used, the scaling factor in the BN layers are subtracted by $1$ at initialization. This modification ensures the BN layers are not changed initially. We find it important to fix the scaling factor term in BN layer during training or the DCM could learn trivial solutions. 

\textbf{Testing Procedure}
As discussed in Section \ref{sec:inference}, we use random as well as the first few images from the same video clip as input to the DCM model. For random sampling the procedure we report both the mean and the std of the mIOU from 20 repetitions. When rejection sampling with entropy is used, the number of DCM evaluations is dynamic. In those cases, we also report the average number of evaluations and computational complexity (in number of Multiply-Adds). Although both testing procedures are evaluated, we report the result from random shuffling by default as it results in a larger test set, making the accuracy numbers more robust. However, our ablation study shows that as expected, there is no significant difference in the two testing procedures.


\subsection{Comparison with Baselines}

\textbf{AlexNet FCN32s}
AlexNet is used to build baseline methods on CamVid segmentation \cite{Alpher16}. In this set of experiments, we use a DCM model based on ResNet-50 backbones. The prediction network is AlexNet FCN32s. The DCM model is always initialized from weights pretrained on ImageNet. For our experiments using the entropy trick, the threshold for rejecting a DCM input is set to $0.1$. 

Table \ref{tab:a} summarizes our results. We carefully compare against best available baseline results reported in the literature. For pretrained AlexNet FCN32s model on $720\times 960$ images, our baseline is comparable to the reported number in \cite{Alpher16}. When training from scratch, our baseline is better than their reported number in the corresponding setting. The proposed method outperforms the baselines by 2-3\% with negligible increase in average complexity in all settings. Large gains are observed for both small picture input as well as large picture input. Although we use ImageNet pretrained models in DCM, we observe that the gain is similar regardless of the initialization method used in the AlexNet backbone. This suggests that the gains are not the result of feature transfer from ImageNet. 

In this experimental setting, the quality of calibration features extracted by DCM is high. As long as the DCM input is from the correct domain, the modulation process almost always leads to improvements in testing accuracy. For this reason, the entropy trick for DCM input selection is not effective in improving the testing accuracy. This is in contrast in our findings from using MobileNet V2 backbones, in which case the entropy brings in significant gains.


\textbf{MobileNet FCN32s}
To further evaluate our method in mobile applications, we experiment with models using MobileNet V2 \cite{Alpher14} as backbones. MobileNet V2 is a recently proposed light-weight architecture designed for mobile applications. It significantly reduces model complexity by adoption of depth-wise convolution and inverted residual units. In our experiments, we use a MobileNet V2-1.0 model as the backbone network of DCM. The prediction networks at FCN32s models using MobileNet V2 models with different width factors. The DCM model is initialized using weights pretrained on CamVid, while the prediction network model uses weights pretrained on ImageNet. 

Table \ref{tab:b} compares our algorithm against baselines. Notably, MobileNet V2 baseline models out-performs AlexNet models despite its much smaller complexity. Our approach further improves the mIOU by 1-2\% with insignificant increase in complexity. The gain seems to increase when the backbone network has smaller width. In particular, the gain using $\{1.0, 0.75, 0.5, 0.35\}$ are 1.0\%, 1.4\%, 2.5\% and 1.2\% respectively. Intuitively, as the prediction model becomes smaller, it should be more difficult for the model to generalize to different domains, thus the benefit from using the DCM model is larger. Our result suggests that this is the general trend. However, when the model is too small (in the case of MobileNet V2-0.35), the benefit starts to diminish. The authors conjectures that when the prediction network is too small, our channel-wise feature modulation mechanism is not powerful enough to significantly improve the feature quality. 

Another interesting observation for is that in experiments using MobileNet V2 backbones, the standard deviations in the prediction accuracy is larger than those in AlexNet. In particular, the standard deviation in prediction accuracy reaches $1.92\%$ for the model based on MobileNet V2-0.35. This is likely caused by the fact that MobileNet V2-0.35 is a smaller architecture. Thus, an incorrect modulation signal from DCM could have a greater impact to its prediction accuracy. When sub-optimal DCM inputs are used, modulating the prediction networks leads to accuracy numbers that are below the baselines. In this context, we find that high entropy of the domain classifier ($C_D$ in Figure \ref{fig:pipeline}) can reliably indicate a non-ideal input to DCM. We test the rejection sampling procedure using an entropy threshold of $0.05$. This improves accuracy in all settings, but the gain for MobileNet V2-0.35 is a particularly large at $1\%$. It also leads to a more predictable algorithm at test time as can be concluded from the largely reduced standard deviation in mIOU numbers.

\subsection{Ablation Studies}

To further understand the proposed method, we use ablation studies to answer the following questions. 

\begin{table}[]
\footnotesize
\begin{center}
\begin{tabular}{l|cc|cl}
\hline
Backbone & Img. Size & Pretrained & Eval Img. & mIOU($\%$) \\ \hline
AlexNet & $360\times 480$ & $\times$ & $4(-0.55)$ & $44.7(+0.3)$ \\
AlexNet & $360\times 480$ & \checkmark & $3(-0.55)$ & $53.1(+0.0)$ \\
AlexNet & $720\times 960$ & $\times$ & $2(-1.75)$ & $50.8(-1.0)$ \\
AlexNet & $720\times 960$ & \checkmark & $3(-0.1)$ & $61.1(+0.2)$ \\ \hline
MobileV2-1.0 & $360\times 480$ & \checkmark & $3(+0.55)$ & $59.5(-0.1)$ \\
MobileV2-0.75 & $360\times 480$ & \checkmark & $2(-0.35)$ & $59.6(+0.1)$ \\
MobileV2-0.5  & $360\times 480$ & \checkmark & $3(-0.15)$ & $56.9(+0.0)$ \\
MobileV2-0.35 & $360\times 480$ & \checkmark & $3(+0.5)$ & $52.9(-0.1)$ \\ \hline
\end{tabular}
\caption{Use starting frames in a video instead of randomly selected images for DCM input. The numbers enclosed in parenthesis show the relative change in numbers against the random selection counterparts. Eval Img. denotes to number of images the DCM processed.}
\label{tab:f}
\end{center}
\end{table}

\begin{table}[]
\footnotesize
\begin{center}
\begin{tabular}{l|cc|cl}
\hline
Backbone & Img. Size & Pretrained & mIOU($\%$) & $\Delta(\%)$ \\ 
\hline
AlexNet & $360\times 480$ & $\times$ & $37.1(\pm 1.31)$ & $-7.3$ \\
AlexNet & $360\times 480$ & \checkmark & $51.2(\pm 0.21)$ & $-1.9$ \\
AlexNet & $720\times 960$ & $\times$ & $41.8(\pm 1.30)$ & $-10.0$ \\
AlexNet & $720\times 960$ & \checkmark & $59.6(\pm 0.31)$ & $-1.3$ \\
\hline
MobileV2-1.0 & $360\times 480$ & \checkmark & $54.4(\pm 1.05)$ & $-5.2$ \\
MobileV2-0.75 & $360\times 480$ & \checkmark & $52.6(\pm 1.20)$ & $-6.9$ \\
MobileV2-0.5 & $360\times 480$ & \checkmark & $50.2(\pm 0.72)$ & $-6.7$ \\
MobileV2-0.35 & $360\times 480$  &\checkmark & $44.3(\pm 1.23)$ & $-8.7$ \\ 
\hline
\end{tabular}
\caption{Effects of inputs with wrong domains. $\Delta$ denotes to the difference between ablation study result and our method with using backbone Network+DCM+entropy.}
\label{tab:d}
\end{center}
\end{table}

\textbf{Can DCM leads to performance gain if the input are the first few frames? } To answer this question, we compare the two types of inputs (first frames versus random frames). Table \ref{tab:f} summarizes the comparison. It is clear that the choice of input type does not result in significantly different accuracy or complexity. 

\textbf{What happens if the DCM input is from the wrong domain? } This is an interesting sanity check: If wrong inputs to the DCM does not lead to significant reduction in accuracy, then DCM might be learning producing trivial modulation signals. From Table \ref{tab:d}, using input images with wrong domain ids results in up to 10\% drop in mIOU, dispriving this possibility. 

\textbf{What happens if we add additional scaling parameters on AlexNet?} Since AlexNet does not have BN layers, our modulation mechanism effectively adds new parameters to the network. To rule out a trivial gain resultant from added parameters, we perform a comparison. As can be seen in Table \ref{tab:c}, adding parameters alone does not improve accuracy significantly.

\textbf{What happens if there is domain mismatches in training? } It is reasonable to suspect that the system is simply benefiting from added parameters from the DCM model (although the inputs are out-of-sync to the current frame). To rule out this trivial case, we purposely create mismatches in domain ids between the input to DCM and the prediction network. Our experiment summarized in \ref{tab:e} shows that our method significantly outperforms models trained with domain mismatches, disproving another trivial case. 

The answer to the last three questions strongly suggests that our method is indeed using the domain information in a non-trivial manner. Based on these results, we conclude that: (1) Gains from our approach are not a trivial result of naively adding parameters. (2) Nor does it result from learning the degenerate solutions such as dataset means.

\begin{table}[]
\footnotesize
\begin{center}
\begin{tabular}{l|cc|cc}
\hline
Backbone & Img. Size & Pretrained & mIOU($\%$) & $\Delta(\%)$\\ \hline
AlexNet+param & $360\times 480$ & $\times$ & $42.3$ & $-2.2$\\
AlexNet+param & $360\times 480$ & \checkmark & $49.8$ & $-3.3$ \\
AlexNet+param & $720\times 960$ & $\times$ & $49.5$ & $-2.3$ \\
AlexNet+param & $720\times 960$ & \checkmark & $55.2$ & $-5.7$ \\ 
\hline
\end{tabular}
\caption{Adding channel-wise multiplication parameters in the AlexNet. $\Delta$ denotes to the difference between ablation study result and our method with AlexNet+DCM+entropy.}
\label{tab:c}
\end{center}
\end{table}

\begin{table}[]
\footnotesize
\begin{center}
\begin{tabular}{l|cc|lc}
\hline
Backbone & Img. Size & Pretrained & mIOU($\%$) & $\Delta(\%)$ \\ \hline
AlexNet & $360\times 480$ & $\times$ & $43.3(\pm 0.05)$ & $-1.1$\\
AlexNet & $360\times 480$ & \checkmark & $51.8(\pm 0.02)$ & $-1.3$\\
\hline
MobileV2-1.0 & $360\times 480$ & \checkmark & $59.7(\pm 0.16)$ & $+0.1$\\
MobileV2-0.75 & $360\times 480$ & \checkmark & $58.1(\pm 0.05)$ & $-1.4$\\
MobileV2-0.5  & $360\times 480$ & \checkmark & $55.1(\pm 0.10)$ & $-1.8$\\
MobileV2-0.35 & $360\times 480$ & \checkmark & $52.2(\pm 0.10)$ & $-0.8$\\ \hline
\end{tabular}
\caption{Effect of domain mismatches in DCM training. $\Delta$ denotes to the difference between ablation study result and our method with using backbone Network+DCM+entropy.}
\label{tab:e}
\end{center}
\end{table}

%% file: tex/Conclusion.tex
\section{Conclusion and Future Directions}
\label{sec:conclusion}

In this work, we propose and empirically investigate a novel dual frame-rate architecture for efficient video understanding. This strategy has demonstrated consistent gains over baselines, over a wide variety of settings. Through ablation studies, we show that the success is due to accurate modeling of the environment. We also propose practical solutions to improve the robustness of our algorithm when the environmental modeling is inaccurate. 

The current work is limited by the size of the dataset used for evaluation. An important future work is to curate a large-scale dataset to evaluate similar design principles. It is also interesting to test its applicability to a wider variety of applications. In this work we use semantic segmentation on video clips as an example application. The authors expect the same strategy would lead to improvements in related applications such as object detection and instance level segmentation, as the problem structures and constraints are similar. Another interesting direction is to find the ``optimal'' modulation strategy and a potential generalization of the two-stream design. In fact, temporally visual signals usually exhibits multi-scale structures, just as they do spatially. It would be interesting to go beyond the hand-crafted dual frame-rate design to a truly adaptive multi frame-rate system. 